\documentclass{article}

\usepackage[final]{neurips_2020_tda}
\usepackage[utf8]{inputenc} % allow utf-8 input
\usepackage[T1]{fontenc}    % use 8-bit T1 fonts
\usepackage{amsfonts}       % blackboard math symbols
\usepackage{booktabs}       % professional-quality tables
\usepackage{hyperref}       % hyperlinks
\usepackage{url}            % simple URL typesetting
\usepackage{microtype}      % microtypography
\usepackage{nicefrac}       % compact symbols for 1/2, etc.
\usepackage{paralist}       % in-paragraph enumerations
\usepackage{siunitx}        % SI units

\usepackage{graphicx}
\usepackage{booktabs}
\usepackage{multirow}
\usepackage{graphicx}
\usepackage{pifont}

\title{\textit{giotto-tda}: A Topological Data Analysis Toolkit\\ for Machine Learning and Data Exploration}

\author{Guillaume Tauzin$^{1, \, 2}$ \\ gtauzin@protonmail.com \\
  \And
  Umberto Lupo$^{3, 4}$ \\ umberto.lupo@epfl.ch \\
  \And
  Lewis Tunstall$^4$ \\ lewis.c.tunstall@gmail.com \\
  \And
  Julian Burella Pérez$^5$ \\ julian.burellaperez@heig-vd.ch \\
  \And
  Matteo Caorsi$^4$ \\ m.caorsi@l2f.ch \\
  \And
  Wojciech Reise$^6$ \\ reisewojciech@gmail.com \\
  \And
  Anibal M. Medina-Mardones$^2$ \\ anibal.medinamardones@epfl.ch \\
  \And
  Alberto Dassatti$^5$ \\ alberto.dassatti@heig-vd.ch \\
  \And
  Kathryn Hess$^2$ \\ kathryn.hess@epfl.ch \\
  \And \\
  $^1$INAIT SA \\
  $^2$Laboratory for Topology and Neuroscience, EPFL \\
  $^3$Laboratory of Computational Biology and Theoretical Biophysics, EPFL \\
  $^4$L2F SA \\
  $^5$School of Management and Engineering Vaud, HES-SO, \protect\\ University of Applied Sciences Western Switzerland \\
  $^6$ DataShape, Inria Saclay -- \^{I}le-de-France\\
}

\setcitestyle{square}
\setcitestyle{numbers}
\begin{document}

\maketitle

\begin{abstract}
  We introduce \emph{giotto-tda}, a \texttt{Python} library that integrates high-performance topological data analysis with machine learning via a \emph{scikit-learn}--compatible API and state-of-the-art \texttt{C++} implementations. The library's ability to handle various types of data is rooted in a wide range of preprocessing techniques, and its strong focus on data exploration and interpretability is aided by an intuitive plotting API. Source code, binaries, examples, and documentation can be found at \url{https://github.com/giotto-ai/giotto-tda}.
\end{abstract}

\section{Introduction}

Topological Data Analysis (TDA) uses tools from algebraic and combinatorial topology to extract features that capture the shape of data~\citep{topdata}. In recent years, algorithms based on topology have proven very useful in the study of a wide range of problems. In particular, \emph{persistent homology} has had significant impact on data intensive challenges including the classification of porous materials~\citep{materials}, the study of structures in the weight space of CNNs~\citep{cnn}, and the discovery of links between structure and function in the brain~\citep{cavities}. The \emph{Mapper} algorithm has also received considerable attention after its use in the identification of a highly treatable subgroup of breast cancers~\citep{cancer}.

Despite its power and versatility, TDA has remained outside the toolbox of most Machine Learning (ML) practitioners, largely because current implementations are developed for research purposes and not in high-level languages. The aim of \emph{giotto-tda} is to fill this gap by making TDA accessible to the \texttt{Python} data science community, while supporting research. To this end, \emph{giotto-tda} inherits the flexibility of \emph{scikit-learn}, the most popular all-purpose ML framework~\citep{sklearn}, and extends it with TDA capabilities including a wide range of persistent homology and Mapper-type algorithms. It enables TDA to be applied to univariate and multivariate time series, images, graphs, and their higher dimensional analogues, simplicial complexes. This makes \emph{giotto-tda} the most comprehensive \texttt{Python} library for topological \emph{machine learning} and data exploration to date.

\section{Architecture}

To use topological features in machine learning effectively, techniques such as hyperparameter search and feature selection need to be applied at a large scale. Facilitating these processes is one of the reasons why \emph{giotto-tda} maintains and extends compatibility with the \emph{scikit-learn} API. \emph{giotto-tda} provides users with full flexibility in the design of TDA pipelines via modular \texttt{estimators}, and the highly visual nature of topological signatures is harnessed via a plotting API based on \emph{plotly}. This exposes a set of external functions and class methods to plot and interact with intermediate results represented as standard \emph{NumPy} arrays~\citep{numpy}.

To combine TDA methods with the many time-delay embedding techniques used frequently in time series prediction \citep{perea, munch}, one must allow \texttt{transformers} extra flexibility not present in the basic architecture of \emph{scikit-learn}. To support this task, \emph{giotto-tda} provides a novel \texttt{TransformerResamplerMixin} class, as well as an extended version of \emph{scikit-learn}'s \texttt{Pipeline}.~\footnote{The interested reader is referred to \url{https://giotto-ai.github.io/gtda-docs/0.3.1/notebooks/time_series_forecasting.html} for a tutorial on these concepts and features.}

Through \textit{scikit-learn}--based wrapper libraries for  \textit{PyTorch}~\citep{pytorch} such as
\textit{skorch}~\citep{skorch} and the \textit{scikit-learn} interface offered in \textit{TensorFlow}~\citep{tensorflow}, it is also possible to use deep learning models as final estimators in a \textit{giotto-tda} \texttt{Pipeline}.

\section{Persistent homology}

Persistent homology is one of the main tools in TDA. It extracts and summarises, in so-called persistence diagrams, multi-scale relational information in a manner similar to hierarchical clustering, but also considering higher-order connectivity. It is a very powerful and versatile technique. To fully take advantage of it in ML and data exploration tasks, \emph{giotto-tda} offers \emph{scikit-learn}--compatible components that enable the user to a) transform a wide variety of data input types into forms suitable for computing persistent homology, b) compute persistence diagrams according to a large selection of algorithms, and c) extract a rich set of features from persistence diagrams. The result is a framework for constructing end-to-end \texttt{Pipeline} objects to generate carefully crafted topological features from each sample in an input raw data collection. At a more technical level, features are often extracted from persistence diagrams by first representing them as curves or images, or by defining kernels. Each method for doing so typically comes with a set of hyperparameters that must be tuned to the problem at hand. \emph{giotto-tda} exposes a large selection of such algorithms and, by tightly integrating with the \emph{scikit-learn} API for hyperparameter search, cross-validation and feature selection, allows for simple data-driven tuning of the many hyperparameters involved.

\begin{figure}[h!]
  \begin{scriptsize}
    \centering
    \includegraphics[width=0.90\linewidth]{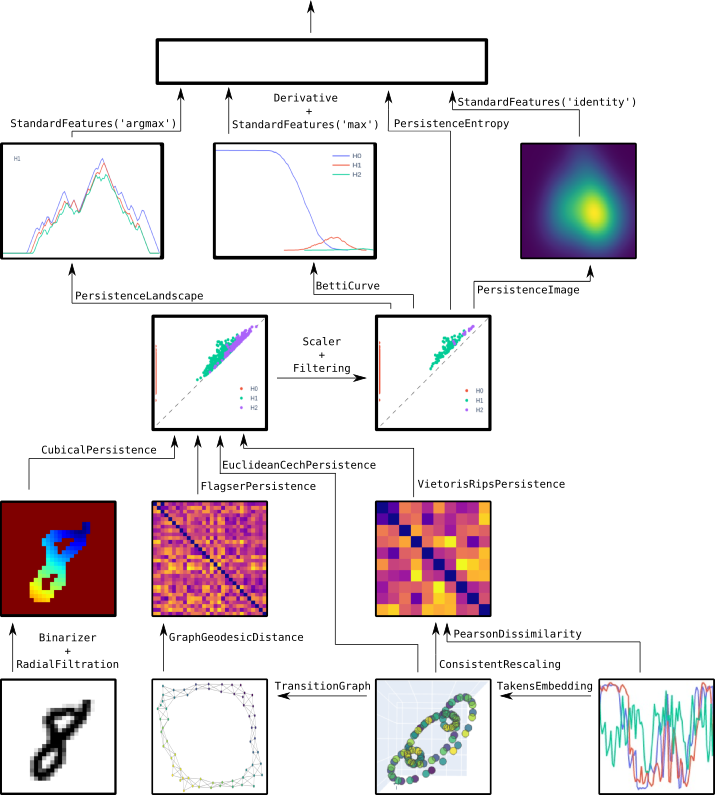}
    \put(-199,386){\scriptsize{\textit{scikit-learn} \texttt{Estimator}}}
    \put(-378,356){ \rotatebox{90}{\footnotesize{Feature}}}
    \put(-255,365){{$\left[ 32.56, 567.42, \dots, 906.08, 23.09 \right]$}}
    \put(-113,356){ \rotatebox{90}{\scriptsize{Vector}}}
    \put(-38,268){ \rotatebox{90}{\scriptsize{Image}}}
    \put(-168,268){ \rotatebox{90}{\scriptsize{Curve}}}
    \put(-275,268){ \rotatebox{90}{\scriptsize{Curve}}}
    \put(-378,268){ \rotatebox{90}{\footnotesize{Representation}}}
    \put(-223,182){ \rotatebox{90}{\scriptsize{Diagram}}}
    \put(-111,182){ \rotatebox{90}{\scriptsize{Diagram}}}
    \put(-378,182){ \rotatebox{90}{\footnotesize{Pers.\ diagram}}}
    \put(-111,90){ \rotatebox{90}{\scriptsize{Distance}}}
    \put(-103,90){ \rotatebox{90}{\scriptsize{matrix}}}
    \put(-223,90){ \rotatebox{90}{\scriptsize{Weighted adjacency}}}
    \put(-215,90){ \rotatebox{90}{\scriptsize{matrix}}}
    \put(-299,90){ \rotatebox{90}{\scriptsize{Image}}}
    \put(-378,100){ \rotatebox{90}{\footnotesize{Prepr.\ data}}}
    \put(-299,1){ \rotatebox{90}{\scriptsize{Image}}}
    \put(-223,1){ \rotatebox{90}{\scriptsize{Graph}}}
    \put(-111,1){ \rotatebox{90}{\scriptsize{Point cloud}}}
    \put(0,1){ \rotatebox{90}{\scriptsize{Time series}}}
    \put(-378,11){ \rotatebox{90}{\footnotesize{Raw data}}}
    \caption{Non-exhaustive depiction of \textit{giotto-tda} capabilities. Arrows represent operations available as \texttt{transformers} and paths potential \texttt{pipelines}.}
    \label{diagram}
  \end{scriptsize}
\end{figure}
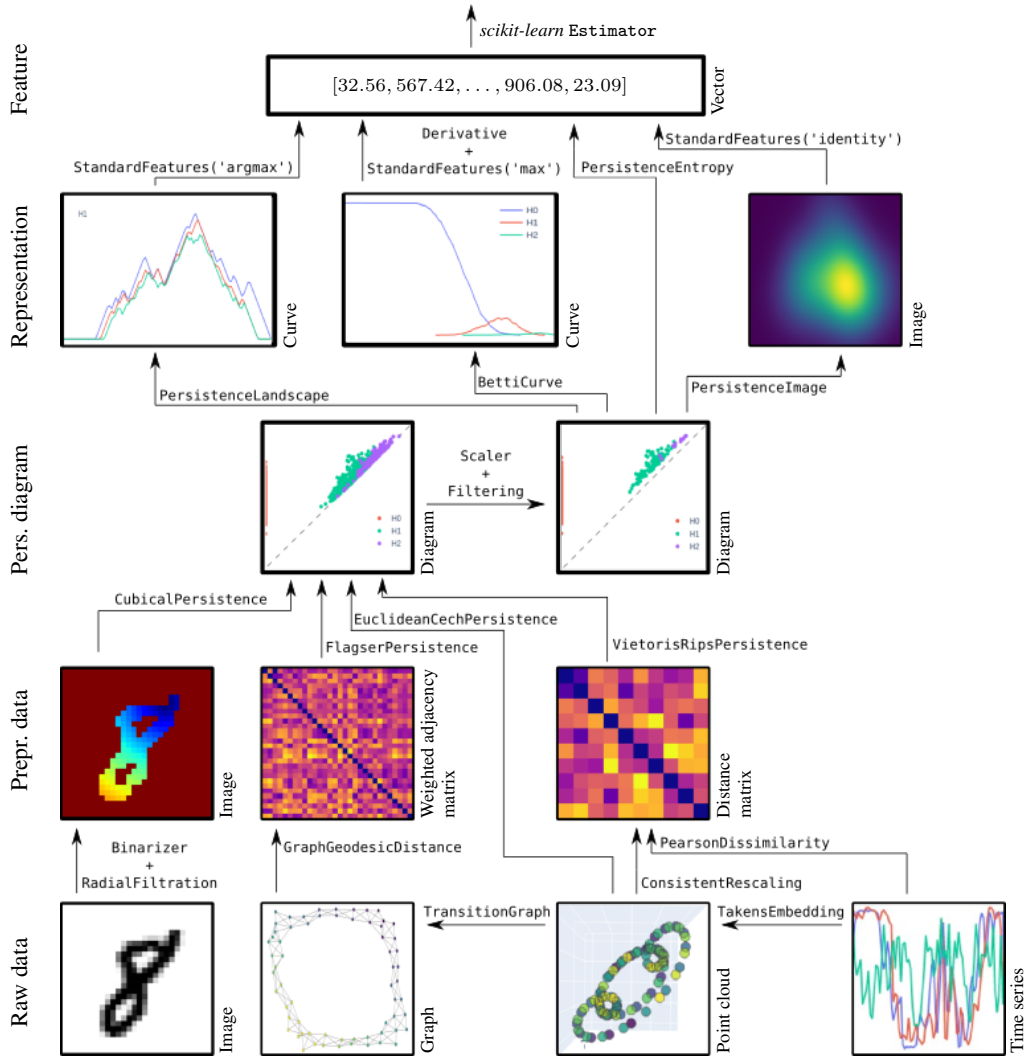

In Figure~\ref{diagram}, we present some of the many possible feature-generation workflows that are made available by \textit{giotto-tda}, starting with a sample in the input raw data collection.

% Guillaume
% The result is a framework for constructing end-to-end \texttt{Pipeline} objects to either carefully design topological features by conducting a hyperparameters search or massively generate them and rely on a feature selection algorithm. In Figure~\ref{diagram}, some of the many possible workflows to generate feature from each sample in an input raw data collection. made available by \textit{giotto-tda} are represented.

% Guillaume ++
% The result is a framework for constructing end-to-end \texttt{Pipeline} objects that create topological features by either; carefully designing them via hyperparameter searches, or straining a massively generated collection of them through feature selection algorithms. In Figure~\ref{diagram}, we present some of the many possible workflows that are made available by \textit{giotto-tda} for this purpose starting with a sample in the input raw data collection.

A comparison between \textit{giotto-tda} and other \texttt{Python} persistent homology libraries is shown in Table~\ref{features}. A highlight of this comparison is the presence of directed persistent homology~\citep{cavities, flagser}, a viewpoint that emphasises the non-symmetric nature of many real-world interactions. \textit{giotto-tda} provides preprocessing \texttt{transformers} to make use of it for a wide range of input data types.

\begin{table}[h!]
  \centering
  \resizebox{\columnwidth}{!}{%
    \begin{tabular}{llcccc} \toprule
      & & \textit{giotto-tda} v0.3.1 & \textit{GUDHI} v3.3.0  & \textit{scikit-tda} & \textit{Dionysus} 2 \\
      \midrule
      \multirow{3}{*}{time series}  & sliding window          & Yes  & -        & -          & -          \\
      & Takens' embedding       & Yes  & Yes      & -          & -          \\
      & Pearson dissimilarity   & Yes  & -        & -          & -          \\
      \midrule
      \multirow{6}{*}{point clouds \& metric spaces} & consistent rescaling~\cite{consistent}    & Yes  & -        & -          & -          \\
      & $k$-nearest neighbors     & Yes    & Yes      & -          & -          \\
      & subsampling& -    & Yes      & -          & -          \\
      & density    & -    & Yes      & -          & -          \\
      & Gromov--Hausdorff distance& -    & -        & Yes        & -          \\
      & distance to measure     & -    & Yes      & -          & -          \\
      \midrule
      \multirow{3}{*}{images}       & binarizer  & Yes  & -        & -          & -          \\
      & image to point cloud    & Yes  & -        & -          & -          \\
      & height filtration       & Yes  & -        & -          & -          \\
      \midrule
      \multirow{2}{*}{graphs}       & transition graph        & Yes  & -        & -          & -          \\
      & geodesic distance       & Yes  & -        & -          & -          \\
      & flag filtrations   & Yes (flagser)  & -        & -          & -          \\

      \midrule
      \multirow{12}{*}{undirected simplicial persistent homology} & Vietoris--Rips           & Yes  & Yes      & Yes        & Yes        \\
      & sparse Rips& Yes  & Yes      & -          & -          \\
      & weighted Rips           & -    & Yes      & -          & -          \\
      & edge collapse \cite{boissonnat2020edge}          & Yes  & \texttt{C++} only & -          & -          \\
      & \v{C}ech       & Yes  & \texttt{C++} only & Yes        & -          \\
      & alpha      & Yes (weak~\cite{gabrielsson2020topology})        & Yes      & Yes        & -          \\
      & witness    & -    & Yes      & -          & -          \\
      & tangential & -    & Yes      & -          & -          \\
      & extended   & -    & Yes      & Yes        & -          \\
      & zigzag     & -    & -        & -          & Yes        \\
      & lower star & -    & Yes      & Yes        & Yes        \\
      \midrule
      \multirow{2}{*}{other persistent homology}        & directed simplicial           & Yes  & -        & -          & -          \\
      &    cubical        & Yes  & Yes      & -          & -          \\
      \midrule
      \multirow{5}{*}{diagram representations}       & persistence landscape    & Yes  & Yes      & -          & -          \\
      & Betti curves            & Yes  & Yes      & -          & -          \\
      & silhouette & Yes  & Yes      & -          & -          \\
      & heat representation~\cite{reininghaus2015stable}     & Yes  & -        & -          & -          \\
      & persistent image        & Yes  & Yes      & Yes        & -          \\
      \midrule
      \multirow{7}{*}{diagram distances and kernels} & bottleneck distance     & Yes  & Yes      & Yes        & Yes        \\
      & Wasserstein distance    & Yes  & Yes      & -          & Yes        \\
      & persistent Fisher~\cite{fisher}       & -    & Yes      & -          & -          \\
      & heat~\cite{reininghaus2015stable}       & Yes  & Yes      & Yes        & -          \\
      & persistent weighted Gaussian~\cite{persweightgauss} & -    & Yes      & -          & -          \\
      & sliced Wasserstein~\cite{slicedWasserstein}      & -    & Yes      & Yes        & -          \\
      & $L^p$ distance between representations & Yes  & -        & -          & -          \\
      \midrule
      \multirow{8}{*}{diagram features} & prominent points        & -    & Yes      & -          & -          \\
      & ATOL~\cite{atol}      & -    & Yes      & -          & -          \\
      & persistence entropy     & Yes  & Yes      & Yes        & -          \\
      & number of points        & Yes  & -        & -          & -          \\
      & complex polynomial~\cite{complexpol}      & Yes  & Yes      & -          & -          \\
      & topological vector~\cite{topovector}      & -    & Yes      & -          & -          \\
      & amplitude  & Yes  & -        & -          & -          \\
      & curve features          & Yes  & -        & -          & -          \\
      \midrule
      \multirow{7}{*}{plotting}     & time series& Yes  & -        & -          & -          \\
      & point cloud& Yes  & -        & -          & -          \\
      & image      & Yes  & -        & -          & -          \\
      & graph      & Yes  & -        & -          & -          \\
      & diagram    & Yes  & Yes      & Yes        & Yes        \\
      & diagram density          & -    & Yes      & -          & Yes        \\
      & representation          & Yes  & -        & -          & -\\
      \bottomrule
      \\
    \end{tabular}
  }
  \caption{Snapshot of the feature support present on the main \texttt{Python} open source libraries with persistent homology capabilities.~\protect\footnotemark}
  \label{features}
\end{table}

Our library matches the code and documentation standards set by \emph{scikit-learn}, and relies on state-of-the-art external \texttt{C++} libraries \citep{gudhi, ripser, hera, flagser} using new performance-oriented bindings based on \emph{pybind11}~\citep{pybind}. In the case of \texttt{ripser}~\citep{ripser}, bindings from \texttt{ripser.py}~\citep{ripserpy} were adapted. In the case of \texttt{flagser}~\citep{flagser}, no \texttt{Python} API was available prior to \textit{giotto-tda}'s sibling project \textit{pyflagser}.\footnote{Source code available at \url{https://github.com/giotto-ai/pyflagser}.} As concerns the computation of Vietoris--Rips barcodes, \emph{giotto-tda} improves on the state-of-the-art runtimes achieved in \cite{boissonnat2020edge} (and now part of \emph{GUDHI}'s C++ codebase) by combining their edge collapse algorithm with \emph{ripser}. Furthermore, the \emph{joblib} package is used throughout to parallelize computations across batches of data. Whenever possible, we contributed with enhancements and bug fixes to some of \textit{giotto-tda}'s \texttt{C++} and \texttt{Python} dependencies.
\addtocounter{footnote}{-1}

\footnotetext{\textit{GUDHI} \citep{gudhi}, \ \textit{scikit-tda} \citep{scikittda}, \ \textit{Dionysus 2} \citep{dionysus}.}

\section{Mapper}
\begin{figure}[!h]
  \vspace*{-0.1cm}
  \centering
  \scriptsize
  \includegraphics[scale=0.24]{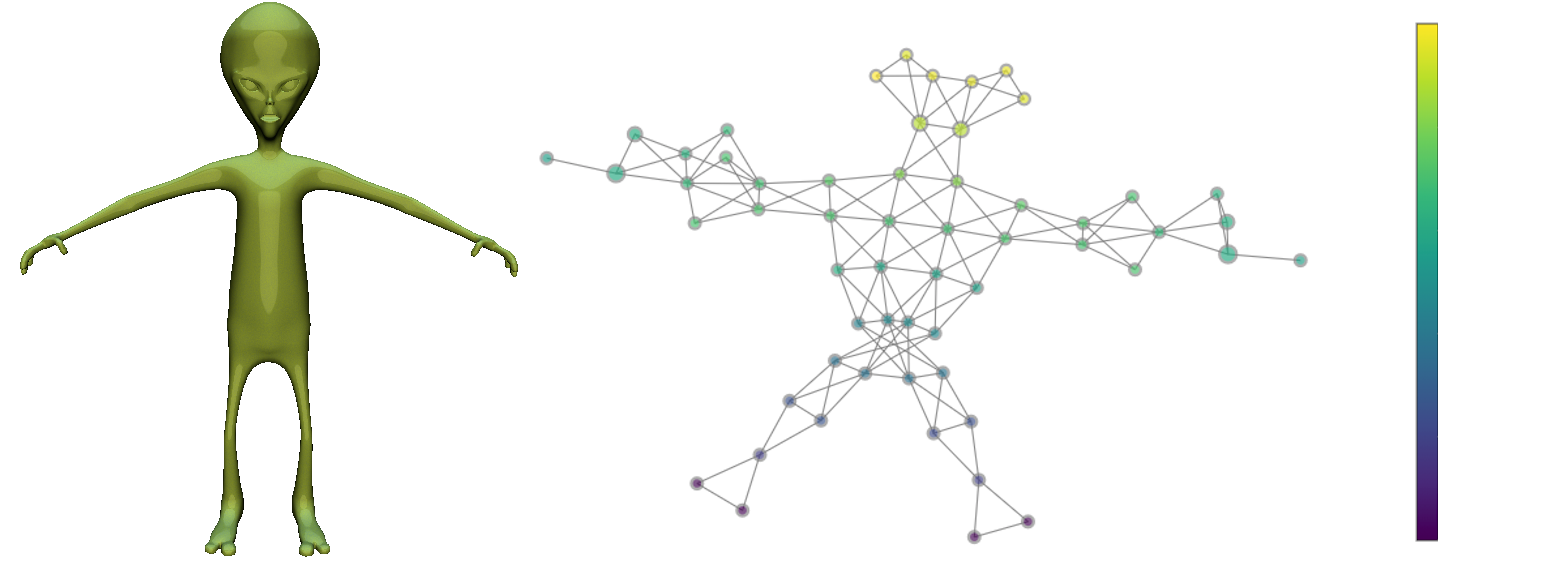}
  \caption[protecting]{Mapper graph generated by \emph{giotto-tda} based on the height of a 3D model.}
  \label{ex_mapper}
  %\vspace*{-0.8cm}
\end{figure}

Mapper is a representation technique of high-dimensional data that, combining the application of filter functions and partial clustering, creates a simple and topologically meaningful description of the input as an unweighted graph (or, more generally, as a simplicial complex). It is primarily used as a data visualization tool to explore substructures of interest in data. In \emph{giotto-tda}, this algorithm is realised as a sequence of steps in a \emph{scikit-learn} \texttt{Pipeline}, where the clustering step can be parallelized. The resulting graph is visualized through an interactive plotting API. This design choice provides a great deal of interoperability and computational efficiency, allowing users to a) realize relevant steps of the Mapper algorithm through any \emph{scikit-learn} \texttt{Estimator}, b) integrate Mapper pipelines as part of a larger ML workflow, and c) make use of memory caching to avoid unnecessary re-computations. Memory caching is especially useful for interactive plotting, where \emph{giotto-tda} allows users to tune Mapper's hyperparameters and observe how the resulting graph changes in real time. An example of a mapper skeletonization adapted from \citep{tda4phys} is shown in Fig.~\ref{ex_mapper}.

To the best of our knowledge, \emph{KeplerMapper}~\citep{kmapper} is the only alternative open-source implementation of Mapper in \texttt{Python} that provides general-purpose functionality. Although \emph{KeplerMapper} also provides the flexibility to use \emph{scikit-learn} \texttt{estimators} to generate Mapper graphs, it does not implement all steps of the algorithm in a single class and is only partially compatible with \emph{scikit-learn} \texttt{pipelines}. Moreover, it does not implement memory caching or provide real-time hyperparameter interactivity in the visualization.

\section{Project management}

\textbf{Easy installation:} Binary packages are available for all major operating systems on the PyPI package repository and can be installed easily by running \verb|python -m pip install -U giotto-tda|. \vspace{1pt} \\
\textbf{Code quality:} The code is unit-tested throughout using \emph{pytest} and \emph{hypothesis} and, as of v0.3.1, test coverage is at $98\%$. The code follows \texttt{PEP8} standards and adheres to the \texttt{Python} coding guideline and \emph{NumPy}-style documentation. CI/CD best practices are in place via Azure Pipelines. \vspace{1pt} \\
\textbf{Community-based development:} We base \emph{giotto-tda}’s development on collaborative tools such as Git, GitHub, and Slack. Contributions are encouraged, and we actively make use of GitHub's issue tracker to provide support and discuss ideas. The library is distributed under the GNU AGPLv3 license. \vspace{1pt} \\
\textbf{Documentation and learning resources:} A detailed API reference is provided using \emph{sphinx}.\footnote{Currently hosted at \url{https://giotto-ai.github.io/gtda-docs/latest/modules/index.html}.} To lower the entry barrier, we provide a theory glossary and a wide range of tutorials and examples that help new users explore how TDA-based ML pipelines can be applied to datasets of various sorts. \vspace{1pt} \\
\textbf{Project relevance:} At the time of writing, the GitHub repository has attracted over 300 stars and between 500 and 1000 visits per week. The PyPI package is downloaded 350 times per month. The library appears in \emph{scikit-learn}'s curated list of related projects.

\section{Concluding remarks}
The very active research field of TDA provides algorithms that can be used at any step of a ML pipeline. \emph{giotto-tda} aims to make these algorithms available in a form that is useful to both the research and data science communities, thus allowing them to use TDA as a part of large-scale ML tasks. We have written \emph{giotto-tda} under the code and documentation standards of \emph{scikit-learn} and, alongside further performance optimization of the existing \texttt{C++} code, future developments will include the first implementation of novel TDA algorithms such as persistence Steenrod diagrams~\citep{steenrod}.

% \newpage

\section*{Acknowledgements}
We thank Roman Yurchak, Philippe Nguyen, and Philipp Weiler for their numerous ideas and contributions. Support from Innosuisse (grant number $32875.1$ lP-ICT) is gratefully acknowledged.

\bibliography{bibliography}
\end{document}